\documentclass{svproc}
\usepackage{url}

\usepackage[utf8]{inputenc} 
\usepackage[T1]{fontenc}    
\usepackage{hyperref,url}
\usepackage{booktabs}       
\usepackage{amsfonts,amsmath,float, upgreek}       
\usepackage{graphicx, multirow, hhline, subcaption, enumitem}
\usepackage[thinlines]{easytable}
\usepackage{tikz}
\usetikzlibrary{arrows.meta, calc}

\newcommand{\E}{\mathbb{E}}
\newcommand{\A}{\mathcal{A}}
\newcommand{\nf}[1]{$\mathsf{#1}$}

\begin{document}
\mainmatter              
\title{Deep Reinforcement Learning for \textsf{FlipIt} Security Game}
\titlerunning{Deep Reinforcement Learning for \textsf{FlipIt} Security Game}

\author{Laura Greige\inst{1} \and Peter Chin\inst{1,2,3}}
\authorrunning{Laura Greige and Peter Chin}

\institute{Boston University, Boston, MA, USA\\
\and
Center for Brains, Minds and Machines, MIT, Cambridge, MA, USA\\
\and
CMSA, Harvard University, Cambridge, MA, USA\\
\email{lgreige@bu.edu, spchin@cs.bu.edu}}

\maketitle              

\begin{abstract}
Reinforcement learning has shown much success in games such as chess, backgammon and Go \cite{bib:rl-in-games,bib:td-gammon,bib:go}. However, in most of these games, agents have full knowledge of the environment at all times. In this paper, we describe a deep learning model in which agents successfully adapt to different classes of opponents and learn the optimal counter-strategy using reinforcement learning in a game under partial observability. We apply our model to \nf{FlipIt} \cite{bib:flipIt}, a two-player security game in which both players, the attacker and the defender, compete for ownership of a shared resource and only receive information on the current state of the game upon making a move. Our model is a deep neural network combined with Q-learning and is trained to maximize the defender's time of ownership of the resource. Despite the noisy information, our model successfully learns a cost-effective counter-strategy outperforming its opponent's strategies and shows the advantages of the use of deep reinforcement learning in game theoretic scenarios. We also extend \nf{FlipIt} to a larger action-spaced game with the introduction of a new lower-cost move and generalize the model to $n$-player \nf{FlipIt}.
\end{abstract}

\section{Introduction}

Game theory has been commonly used for modeling and solving security problems. When payoff matrices are known by all parties, one can solve the game by calculating the Nash equilibria of the game and by playing one of the corresponding mixed strategies to maximize its gain (or symmetrically, minimize its loss). However, the assumption that the payoff is fully known by all players involved is often too strong to effectively model the type of situations that arise in practice. It is therefore useful to consider the case of incomplete information and apply reinforcement learning methods which are better suited to tackle the problem in these settings. In particular, we examine the two-player game \nf{FlipIt} \cite{bib:flipIt} where an attacker and a defender compete over a shared resource and where agents deal with incomplete observability.

The principal motivation for the game \nf{FlipIt} is the rise of Advanced Persistent Threats (APT) \cite{bib:stuxnet,bib:rsa}. APTs are stealthy and constant computer hacking processes which can compromise a system or a network security and remain undetected for an extended period of time. Such threats include intellectual property theft, host takeover and compromised security keys, caused by network infiltration, typically of large enterprises or governmental networks. For host takeover, the goal of the attacker is to compromise the device, while the goal of the defender is to keep the device clean through software reinstallation or through other defensive precautions. We would like to learn effectively how often should the defender clean the machines and when will the attacker launch its next attack. Hence, the problem can be formulated as finding a cost-effective schedule through reinforcement learning. All these applications can be modeled by the two-player game \nf{FlipIt}, in which players, attackers and defenders, vie for control of a shared resource. The resource could be a computing device or a password for example, depending on which APT is being modeled.

In our work, we train our model to estimate the opponent's strategy and to learn the best-response to that strategy. Since these estimations highly depend on the information the model gets throughout the game, the challenge comes from the incomplete and imperfect information received on the state of the game. The goal is for the adaptive agents to adjust their strategies based on their observations and good \nf{FlipIt} strategies will help players implement their optimal cost-effective schedule. In the next sections, we present previous related studies and provide a description of the game framework as well as its variants. We then describe our approach to address the problem of learning in partial observability and our model architecture. We demonstrate successful counter-strategies developed by adaptive agents against basic renewal strategies and compare their performance in the original version of \nf{FlipIt} to one with a larger action-space after introducing a lower-cost move. In the last sections, we generalize our model to multiplayer \nf{FlipIt} and discuss the next steps of our project.

\section{Related Work}

Although game theory models have been greatly applied to solve cybersecurity problems \cite{bib:cyber-gt-2003,bib:cyber-gt-2010,bib:cyber-gt-2012}, studies mainly focused on one-shot attacks of known types. \nf{FlipIt} is the first model that characterizes the persistent and stealthy properties of APTs and was first introduced by van Dijk et al \cite{bib:flipIt}. In their paper, they analyze multiple instances of the game with non-adaptive strategies and show the dominance of certain distributions against stealthy opponents. They also show that the Greedy strategy is dominant over different distributions, such as periodic and exponential distributions, but is not necessarily optimal. Different variants and extensions of the game have also been analyzed; these include games with additional ``insider'' players trading information to the attacker for monetary gains \cite{bib:flipIt-insider-1,bib:flipIt-insider-2}, games with multiple resources \cite{bib:flipIt-res} and games with different move types \cite{bib:flipIt-moves}. In all these variants, only non-adaptive strategies have been considered and this limits the analysis of the game framework. Laszka et al. \cite{bib:flipIt-adaptive,bib:flipIt-adaptive-2} proposed a study of adaptive strategies in \nf{FlipIt}, but this was done in a variant of the game where the defender's moves are non-stealthy and non-instantaneous. Oakley et al. \cite{bib:qflip} were the first to design adaptive strategies with the use of temporal difference reinforcement learning in 2-player \nf{FlipIt}.

Machine Learning (ML) has been commonly used in different cybersecurity problems such as fraud and malware detection \cite{bib:ml-fraud,bib:ml-malware}, data-privacy protection \cite{bib:ml-data-privacy} and cyber-physical attacks \cite{bib:ml-cyber-physical}. It has allowed the improvement of attacking strategies that can overcome defensive ones, and vice-versa, it has allowed the development of better and more robust defending strategies in order to prevent or minimize the impact of these attacks. Reinforcement Learning (RL) is a particular branch in ML in which an agent interacts with an environment and learns from its own past experience through exploration and exploitation without any prior or with limited knowledge of the environment. RL and the development of deep learning have lead to the introduction of Deep Q-Networks (DQNs) to solve larger and more complex games. DQNs were firstly introduced by Mnih et al. \cite{bib:atari} and have since been commonly used for solving games such as backgammon, the game of Go and Atari \cite{bib:go,bib:go-tree,bib:atari-2}. They combine deep learning and Q-learning \cite{bib:rl-intro,bib:q-learn} and are trained to learn the best action to perform in a particular state in terms of producing the maximum future cumulative reward. Hence, with the ability of modeling autonomous agents that are capable of making optimal sequential decisions, DQNs represent the perfect model to use in an adversarial environment such as \nf{FlipIt}. Our paper extends the research made in stealthy security games with the introduction of adaptive DQN-based strategies allowing agents to learn a cost-effective schedule for defensive precautions in \nf{FlipIt} and its variants, all in real-time.

\section{Game Environment}

\subsection{Framework}

\nf{FlipIt} is an infinitely repeated game where the same one-shot stage game is played repeatedly over a number of discrete time periods. At each period of a game, players decide what action to take depending on their respective strategies. They take control of the resource by moving, or by what is called ``flipping''. Flipping is the only move option available and each player can flip at any time throughout the game. We assume that the defender is the rightful owner of the resource and as such, ties are broken by assigning ownership to the defender. Each player pays a certain move cost for each flip and is rewarded for time in possession of the resource. For our purpose we have used the same reward and flip cost for all players (attackers and defenders, adaptive and non-adaptive), but our environment can be easily generalized in order to experiment with different rewards and costs for both players. Moreover, an interesting aspect in \nf{FlipIt} is that contrary to games like Backgammon and Go, agents do not take turn moving. A move can be made at any time throughout the game and therefore a player's score highly depends on its opponent's moves. The final payoff corresponds to the sum of the player's payoffs from each round. Finally, players have incomplete information about the game as they only find out about its current state once they flip. In particular, adaptive agents only receive feedback from the environment upon flipping, which corresponds to their opponent's last move (LM).

Unless stated otherwise, we assume in the remainder of the paper that the defender is the initial owner of the resource, as it usually is the case with security keys and other devices. The defender is considered to be LM playing a DQN-based strategy against an attacker that follows one of the renewal strategies we describe in the following sections.

\subsection{Markov Decision Process}

Our environment is defined as a Markov Decision Process (MDP). At each iteration, agents select an action from the set of possible actions $\A$. In \nf{FlipIt}, the action space is restrained to two actions: to flip and not to flip. As previously mentioned, agents do not always have a correct perception of the current state of the game. In Figure \ref{fig:state_update} we describe a case where an LM agent $P_1$ plays with only partial observability against a periodic agent $P_2$. When $P_1$ flips at iteration 6, the only feedback it receives from the environment concerns its opponent's flip at iteration 4. Hence, no information is given regarding the opponent's previous flip at iteration 2 and $P_1$ is subjected to an incorrect assumption on the time it controlled the resource. Suppose $P_1$ claims ownership of the resource at iteration 6. Then, $P_1$'s benefit would be equal to the sum of the operational cost of flipping and the reward for being in control of the resource, which is represented by $\uptau_{P_1}$ in the figure below.

\begin{figure}[H]
\centering
\begin{tikzpicture}[scale=0.7]
\node[text width=1.5cm, color=blue] at (-1.4,0.4) {player $P_1$};
\node[text width=1.5cm, color=red] at (-1.4,-0.4) {player $P_2$};
\node[text width=1cm] at (10,-0.4) {};
\draw[fill=red] (0,-1) circle (3pt);
\draw[fill=red] (0,0) rectangle (1,-0.5);
\node[color=black] at (1.5,2) {$\tau_{P_1}$};
\draw[thick,arrows=->] (1,1.5) -- (2,1.5);
\draw[fill=blue] (1,1) circle (3pt);
\draw[fill=blue] (1,0) rectangle (2,0.5);
\draw[fill=red] (2,-1) circle (3pt);
\draw[fill=red] (4,-1) circle (3pt);
\draw[fill=red] (2,0) rectangle (6,-0.5);
\draw[fill=blue] (6,1) circle (3pt);
\draw[fill=red] (6,-1) circle (3pt);
\draw[fill=blue] (6,0) rectangle (8,0.5);
\draw[thick] (0,-0.7) -- (0,0.7);
\draw[thick,arrows=->] (0,0) -- (9,0);
\foreach \x in {0,1,2,...,7}{
\coordinate (A\x) at ($(1,0)+(\x*1cm,0)$) {}; 
\draw[thick] ($(A\x)+(0,0.1)$) -- ($(A\x)-(0,0.1)$); 
}
\end{tikzpicture}
\caption{An example of incomplete or imperfect observability in \nf{FlipIt} by $P_1$}\label{fig:state_update}
\end{figure}

\textbf{State Space}. We consider a discrete state space where each state indicates the current state of the game, i.e. whether the defender is the current owner of the resource and the times elapsed since each agent's last known moves. The current owner of the resource can be inferred from the current state of the game. Agents only learn the current state of the game once they \nf{flip}, causing imperfect information in their observations, as previously explained.

\textbf{Action Space}. In the original version of \nf{FlipIt}, the only move option available is to flip. An adaptive agent therefore has two possible actions: to flip or not to flip. In this paper, we extend the game framework to a larger action-spaced game and introduce a new move called \nf{check}. This move allows an agent to check the current state of the game and obtain information regarding its opponents' last known moves all while paying a lower operational cost than the one of flipping. Just like flipping, an agent can check the state of the game at any time throughout the game and the action spaces are then denoted by $\mathcal A_d = \{\texttt{void}, \texttt{flip}, \texttt{check}\}$ for the defender and $\mathcal A_a = \{\texttt{void}, \texttt{flip}\}$ for the attacker.

\textbf{State Transitions}. Since an agent can move at any time throughout the game, it is possible that both agents involved \nf{flip} simultaneously and ties are broken by automatically assigning ownership to the defender. At each iteration, the state of the game is updated as such. If the defender flipped, the current owner is assigned to the defender. If the defender did not flip and its opponent flipped, the current owner is assigned to the opponent. If neither agent flips, the current owner is left unchanged. The transition to the next step only depends on the current state and actions taken such that the state transition function $T$ is defined by $T : S \times \mathcal A_d \times \mathcal A_a \rightarrow \Delta(S)$.

\textbf{Reward System}. We define the immediate reward at each iteration based on the action taken as well as the owner of the resource at the previous iteration. 

\begin{enumerate}[nosep, wide, labelwidth=!, labelindent=2.4em, label=\roman*)]
	\item \textbf{Operational Costs and Payoff}. Let $r_t$ be the immediate reward received by an agent at time step $t$. We have,

\begin{equation*}
r_t = \begin{cases}
0 &\text{if no play}\\
-~C_c &\text{if $a_t =$ \nf{check}}\\
\uptau \cdot r - C_f &\text{if $a_t =$ \nf{flip}}
\end{cases}
\end{equation*}
where $r$ is the payoff given for owning the resource at one time step, $C_c$ is the operational cost of checking and $C_f$ the operational cost of flipping. $\uptau$ defines the time elapsed between the agent's last flip move and the time step he last owned the resource previous to its current flip, as described in Figure \ref{fig:state_update}.

\item \textbf{Discount Factor}.  Let $\gamma$ be the discount factor. The discount factor determines the importance of future rewards, and in our environment, a correct action at some time step $t$ is not necessarily immediately rewarded. In fact, by having a flip cost higher than a flip reward, an agent is penalized for flipping at the correct moment but is rewarded in future time steps. This is why we set our discount factor $\gamma$ to be as large as possible, giving more importance to future rewards and forcing our agent to aim for long term high rewards instead of short-term ones.
\end{enumerate}

\section{Model Architecture}

Q-learning is a reinforcement learning algorithm in which an agent or a group of agents try to learn the optimal policy from their past experiences and interactions with an environment. These experiences are a sequence of state-action-rewards. In its simplest form, Q-learning is a table of values for each state (row) and action (column) possible in the environment. Given a current state, the algorithm estimates the value in each table cell, corresponding to how good it is to take this action in this particular state. At each iteration, an estimation is repeatedly made in order to improve the estimations. This process continues until the agent arrives to a terminal state in the environment. This becomes quite inefficient when we have a large number or an unknown number of states in an environment such as \nf{FlipIt}. Therefore in these situations, larger and more complex implementations of Q-learning have been introduced, in particular, Deep Q-Networks (DQN).

Deep Q-Networks were firstly introduced by Mnih et al. (2013) and have since been commonly used for solving games. DQNs are trained to learn the best action to perform in a particular state in terms of producing the maximum future cumulative reward and map state-action pairs to rewards. Our objective is to train our agent such that its policy converges to the theoretical optimal policy that maximizes the future discounted rewards. In other words, given a state $s$ we want to find the optimal policy $\pi^*$ that selects action $a$ such that $a = arg\max_{a}\left[~Q_{\pi^*}(s,a)~\right]$ where $Q_{\pi^*}(s,a)$ is the Q-value that corresponds to the overall expected reward, given the state-action pair $(s,a)$. It is defined by,

\begin{align}
	Q_{\pi^*}(s,a) = \E_{\pi}~\Big[~r_t + \gamma r_{t+1} + &\gamma^2 r_{t+2} + ... + \gamma^{T-t} r_{T} \Big|~s_t=s, a_t = a~\Big]
\end{align}\\
where $T$ is the length of the game. Q-values are updated for each state and action using the following Bellman equation,

\begin{align}
	Q_n(s,a) = Q(s,a) + \alpha~\Big[~&R(s,a) + \gamma\max_{a'} Q(s',a') - Q(s,a)~\Big]
\end{align}\\
where $Q_n(s,a)$ and $Q(s,a)$ are the new and current Q-values for the state-action pair $(s,a)$, $R(s,a)$ is the reward received for taking action $a$ at state $s$, $\max_{a'} Q(s',a')$ is the maximum expected future reward given new state $s'$ and all possible actions from state $s'$, $\alpha$ is the learning rate and $\gamma$ the discount factor.

Our model architecture consists of 3 fully connected layers with rectified linear unit (ReLU) activation function at each layer. It is trained with Q-learning using the PyTorch framework \cite{bib:pytorch} and optimized using the Adam optimizer \cite{bib:adam}. We use experience replay \cite{bib:exp-replay} memory to store the history of state transitions and rewards (i.e. experiences) and sample mini-batches from the same experience replay to calculate the Q-values and update our model. The state of the game given as input to the neural network corresponds to the agent's current knowledge on the game, i.e. the time passed since its last move and the time passed since its opponent's last known move. The output corresponds to the Q-values calculated for each action. The learning rate is set to 0.001 while the discount factor is set to 0.99. We value exploration over exploitation and use an $\epsilon$-Greedy algorithm such that at each time step a random action is selected with probability $\epsilon$ and the action corresponding to the highest Q-value is selected with probability $1-\epsilon$. $\epsilon$ is initially set to 0.6 and is gradually reduced at each time step as the agent becomes more confident at estimating Q-values. We choose 0.6 as it yields the best outcome regardless of the attacker's strategy. In particular, we find that, despite eventually converging to its maximal benefit, higher exploration values can negatively impact learning while lower exploration values can cause a slower convergence.

\begin{figure}[H]
\centering
\includegraphics[width=.9\textwidth]{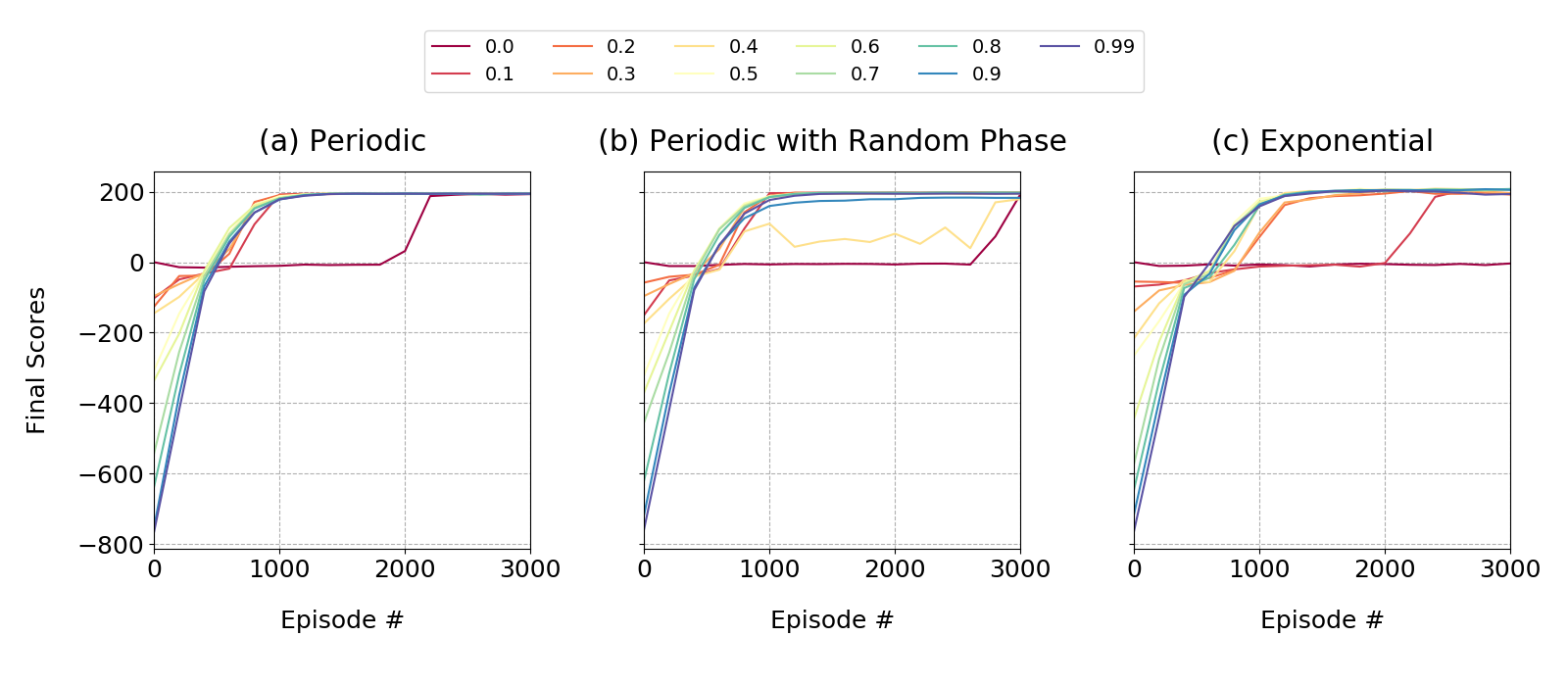}
\caption{Learning overtime averaged over 10 \nf{FlipIt} simulations against 3 renewal strategies, as described in the next section.}
\label{fig:eps}
\end{figure}

\section{Experimental Results}

In what follows, we assume that opponent move rates are such that the expected interval time between two consecutive flips is larger than the flip move cost. We trained our neural network to learn the best counter-strategy to its opponent's such that the reward received for being in control of the resource is set to 1 and the cost of flipping is set to 4. The flip cost is purposely set to a higher value than the reward in order to discourage the defender from flipping at each iteration. The following findings apply for any cost value that is greater than the reward.

\subsection{Renewal Strategies}

A renewal process is a process which selects a renewal time from a probability distribution and repeats at each renewal. For our purpose, a renewal is a flip and our renewal process is 1-dimensional with only a time dimension. There are at least two properties we desire from a good strategy. First, we expect the strategy to have some degree of unpredictability. A predictable strategy will be susceptible to exploitation, in that a malignant or duplicitous opponent can strategically select flips according to the predictable flips of the agent. Second, we expect a good strategy to space its flips efficiently. Intuitively, we can see that near simultaneous flips will waste valuable resources without providing proportionate rewards. In this paper, we examine three basic renewal strategies for the attacker, periodic ($\mathcal P_\delta$), periodic with a random phase ($\mathcal P'_\delta$) and exponential ($\mathcal E_\lambda$), as they present different degrees of predictability and spacing efficiency.

\begin{figure}[H]
\centering
\subcaptionbox{Periodic $\mathcal P_{10}$\label{sub:periodic10}}
{\includegraphics[width=.23\textwidth]{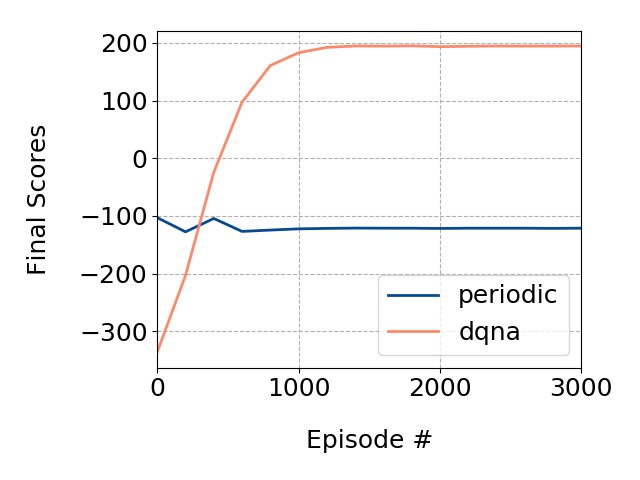}}\quad
\subcaptionbox{Periodic\label{sub:periodic}}
{\includegraphics[width=.23\textwidth]{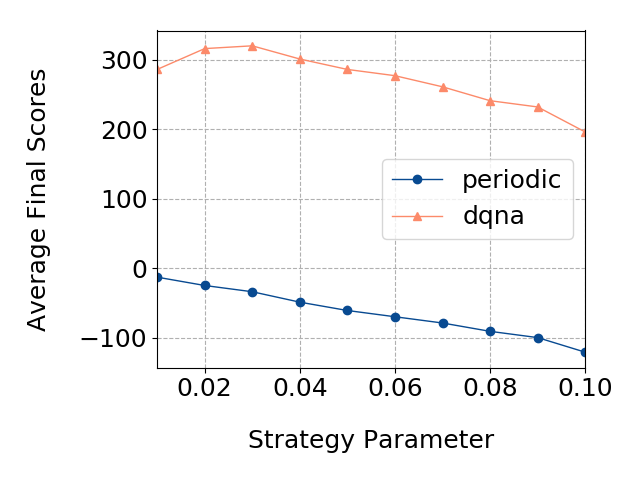}}
\subcaptionbox{PeriodicR\label{sub:periodicR}}
{\includegraphics[width=.23\textwidth]{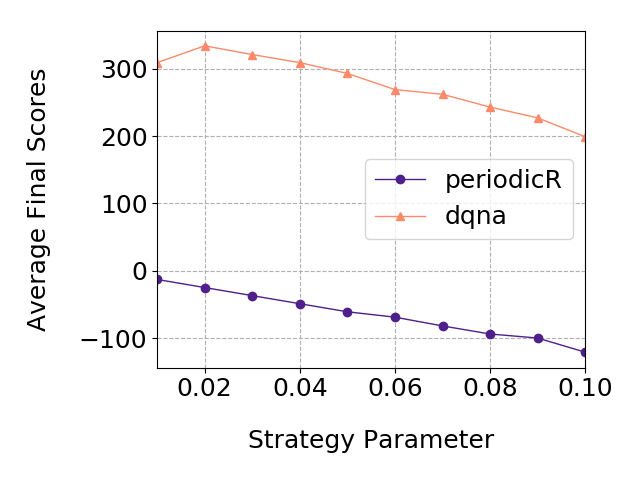}}\quad
\subcaptionbox{Exponential\label{sub:exponential}}
{\includegraphics[width=.23\textwidth]{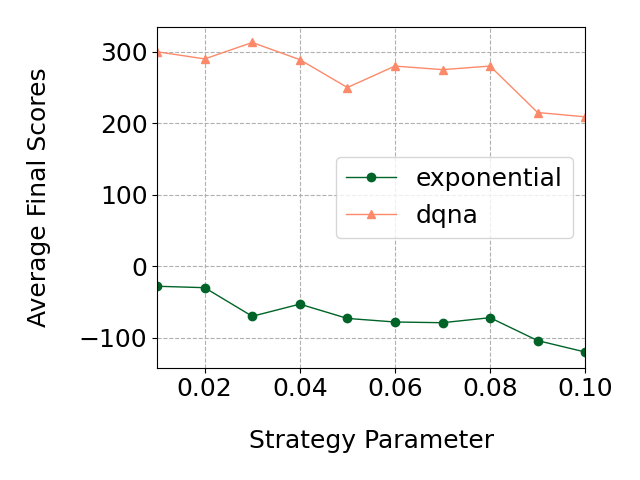}}
\caption{\nf{FlipIt} simulations against renewal strategies over different move rates.}
\label{fig:periodic-example}
\end{figure}

In general, the optimal strategy against any periodic strategy can be found and maximal benefits can be calculated. Since the defender has priority when both players flip simultaneously, the optimal strategy would be to play the same periodic strategy as its opponent's as it maximizes its time of ownership of the resource and reaches maximal benefit. Considering that the cost of flipping is set to 4 and each game is played over 400 iterations, then the theoretical maximal benefit for an adaptive agent playing against a periodic agent with a period $\delta = 10$ would be equal to 200. We oppose an LM adaptive agent to a periodic agent $\mathcal P_{10}$ and show that with each game episode, the defender's final score does in fact converge to its theoretical maximal benefit in Figure \ref{sub:periodic10}.

In Figures \ref{sub:periodic}, \ref{sub:periodicR} and \ref{sub:exponential}, we plot the average final scores after convergence of the defender reward against periodic, periodic with a random phase and exponential strategies, with regards to the opponent strategy parameter. All scores are averaged over 10 runs. In all 3 cases, the defender converges towards its maximal benefit and drives its opponents to negative ones, penalizing them at each action decision. It outperforms all renewal strategies mentioned, regardless of the strategy parameters, and learns the corresponding optimal counter-strategy even against exponential strategies where the spacing between two consecutive flips is random. A more in-depth look into the strategies developed shows that the adaptive agent playing against $\mathcal P_\delta$ and $\mathcal P'_\delta$ learns a strategy where the distribution of wait intervals concentrates on $\delta$ whereas the one playing against $\mathcal E_\lambda$ learns a strategy with a wider spread, spacing its flips efficiently throughout the game. We find that the defender's final score decreases as the attacker move rate increases. This can be explained by the fact that a higher strategy parameter suggests flipping more often and causes the defender to also flip more frequently to counter the attacker, thus causing the overall reward to decrease. Moreover, higher move rates cause shorter interval times between two consecutive flips and this increases the risk of flipping at an incorrect iteration which could penalize the defender; as a matter of fact, in the general case, the worst-case scenario, flipping one iteration before each of its opponents flips, is only one shift away from the optimal strategy, flipping at the same time as the opponents. Despite the decreasing final scores, the defender learns to efficiently counter-attack its opponents, thus maximizing its time of ownership of the resource.

\subsection{Larger Action-Spaced \nf{FlipIt} Extension}

We compare the defender's performance in the original version of \nf{FlipIt} with one where the adaptive agent's action space is extended to $\mathcal A_d = \{\texttt{void}, \texttt{flip}, \texttt{check}\}$. We set the operational cost for checking the current state of the game to 1 as a way to compensate for the benefit of owning the resource at time step $t$ and we run the same experiments against basic renewal strategies in 2-player \nf{FlipIt}.

\begin{figure}[H]
\centering
\includegraphics[width=.3\textwidth]{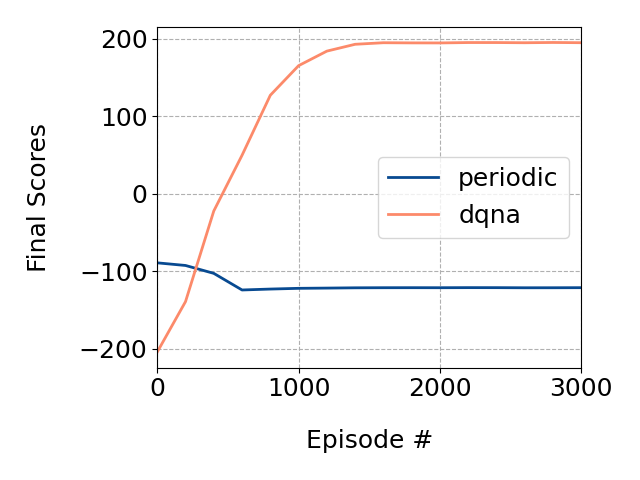}\quad
\includegraphics[width=.3\textwidth]{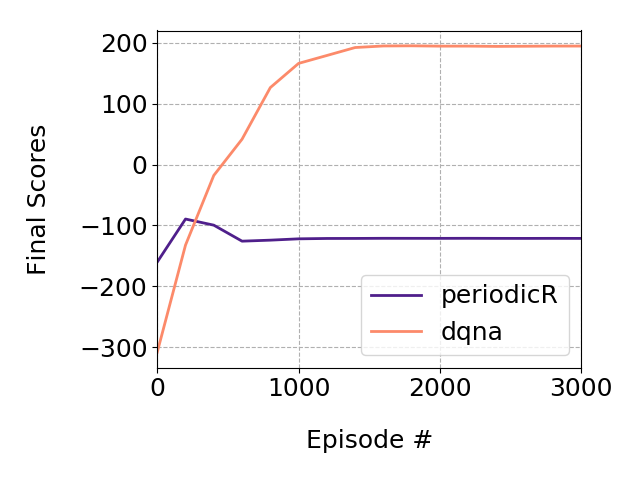}\quad
\includegraphics[width=.3\textwidth]{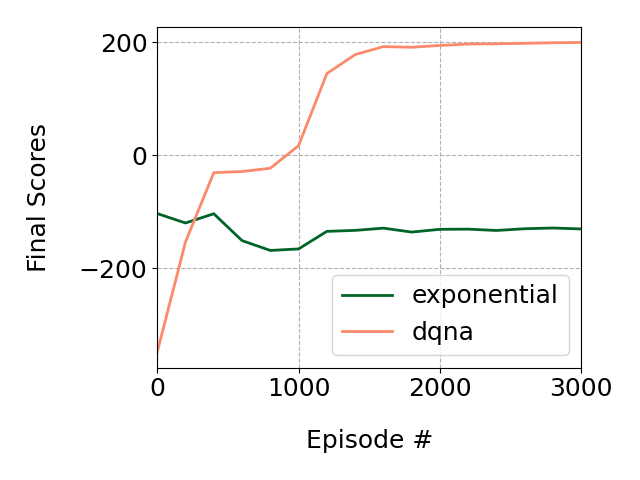}
\subcaptionbox{Periodic\label{sub:periodic10-check}}
{\includegraphics[width=.3\textwidth]{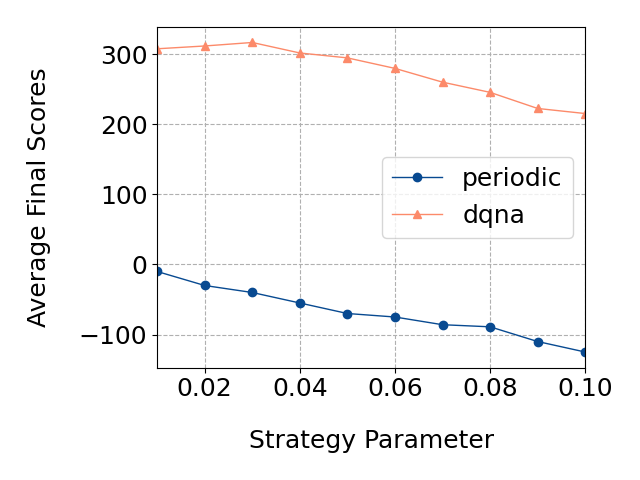}}\quad
\subcaptionbox{PeriodicR\label{sub:periodicR10-check}}
{\includegraphics[width=.3\textwidth]{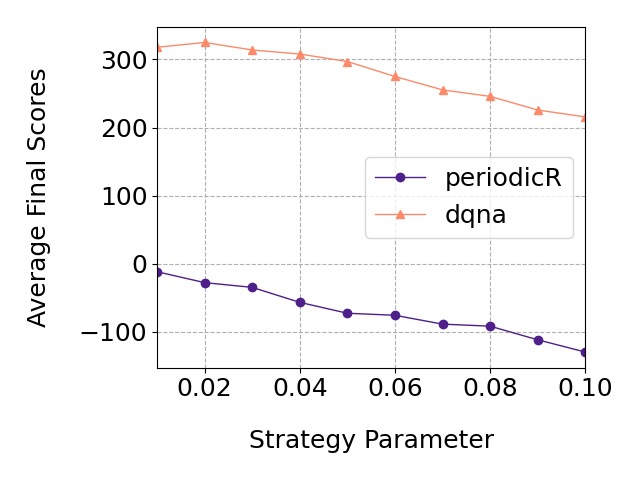}}\quad
\subcaptionbox{Exponential\label{sub:exponential10-check}}
{\includegraphics[width=.3\textwidth]{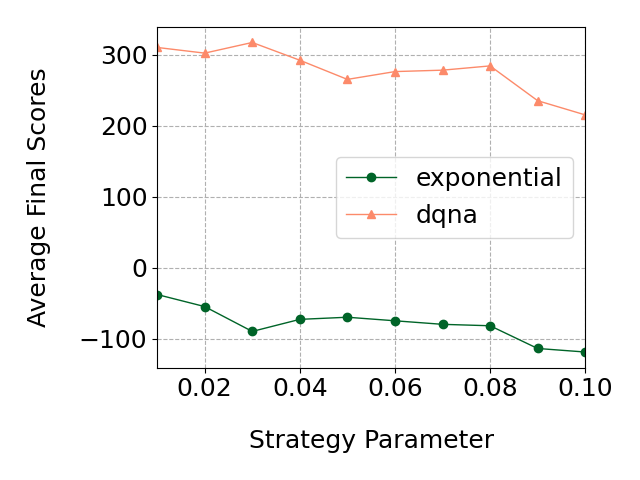}}
\caption{Learning overtime in larger action-spaced \nf{FlipIt}.}
\label{fig:check-example}
\end{figure}

In Figure \ref{fig:check-example}, the top figures represent the DQN's final episode scores against renewal strategies while the bottom ones represent the final scores after convergence of the defender against renewal strategies depending on their move rates, averaged over 10 runs. Here again, the defender reaches maximal benefit against all renewal strategies. Overall, the adaptive agent's learning process is slightly slower than in the original version of the game, but eventually converges to its maximal benefit. In the current setup of \nf{FlipIt}, the addition of \nf{check} might not seem useful and only causes a slower convergence to the defender's maximal benefit. However, when resources are limited and players are only allowed to spend a certain amount on moves, \nf{check} can be key in developing cost-effective strategies, allowing adaptive agents to receive additional feedback from the environment without having to pay an important amount in terms of operational costs.

\subsection{Multiplayer \nf{FlipIt}}

\begin{figure}[h]
  \centering
  \includegraphics[width=\textwidth]{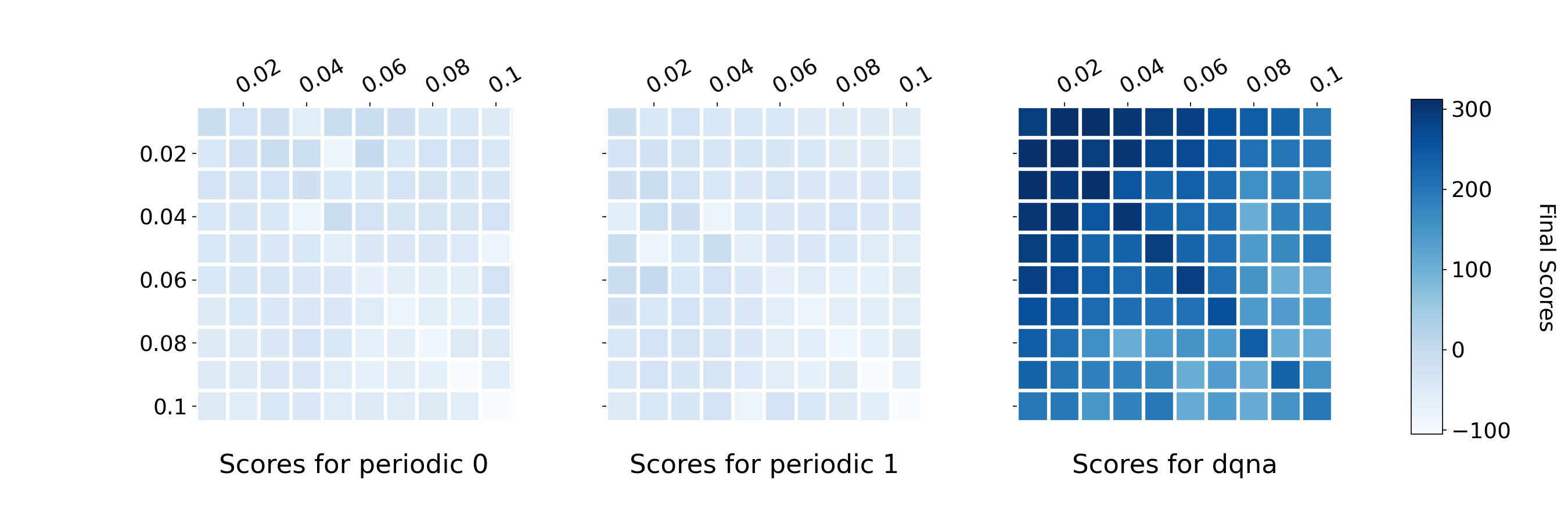}
  \includegraphics[width=\textwidth]{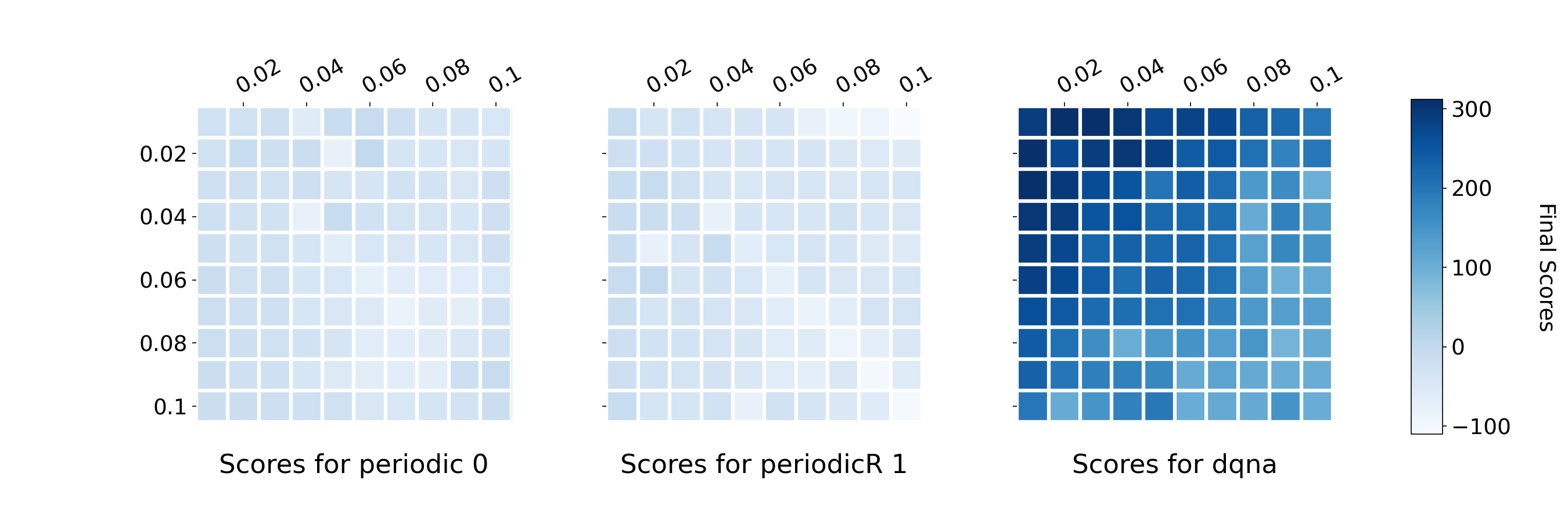}
  \includegraphics[width=\textwidth]{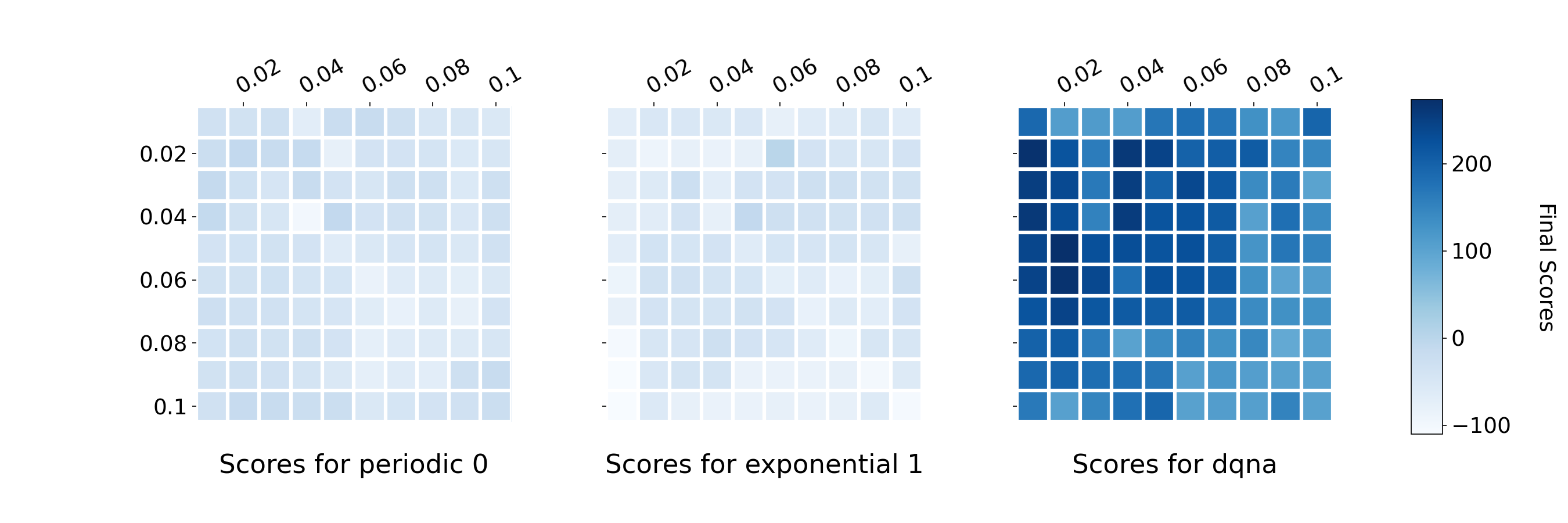}
  \caption{3 Player \nf{FlipIt} simulations}
  \label{fig:nflip-heatmaps}
\end{figure}

Finally, we extend the game to $n$ players, where multiple attackers compete over the shared resource. We assume that one of the attackers is LM ($\textsf{A}_{\textsf{LM}}$) and attempts to adapt its strategy to its opponents' whereas the rest of the players adopt one of the renewal strategies discussed in this paper. As the state of the game corresponds to the agent's knowledge of the game (i.e. opponent's last known moves), the state size increases as $n$ increases. We begin by testing our model by opposing the adaptive agent to a combination of two opponent agents. Players' final scores are averaged over 10 runs are plotted in Figure \ref{fig:nflip-heatmaps}, where darker colors correspond to higher final scores. We show our results with only 3 players for a clearer visualisation of our findings. However, all simulations can be extended to $(n > 3)$-player \nf{FlipIt}.

Consider the case where an adaptive agent plays against two periodic agents with the same move rate $\sigma$. As a reminder, we assume that $\textsf{A}_{\textsf{LM}}$ is the rightful owner of the resource and therefore has priority when assigning a new owner to the resource in the case of simultaneous flips. Therefore, this would be equivalent to playing against one periodic agent with move rate $\sigma$ and we obtain the same results as 2-player \nf{FlipIt}. Now assume both opponents have different move rates. Hypothetically, this scenario would be equivalent to playing against an agent such that its strategy is a combination of both periodic agent strategies. An in-depth look at the strategy learned by $\textsf{A}_{\textsf{LM}}$ shows that the agent learns both strategy periods and spaces its flips accordingly. When opposed against a periodic agent $\mathcal P_\delta$ and an exponential agent $\mathcal E_\lambda$, $\textsf{A}_{\textsf{LM}}$ develops a strategy such that each two flips are efficiently spaced throughout the game, allowing the adaptive agent to converge towards its maximal benefit. As is the case in $2$-player \nf{FlipIt}, the higher the move rates, the smaller the intervals between two consecutive flips are which drives $\textsf{A}_{\textsf{LM}}$ to flip more frequently and causes lower overall final scores. Nonetheless, $\textsf{A}_{\textsf{LM}}$ yields maximal benefit, regardless of its opponents and opponent move rates. 

\subsection{Future Work}

The ultimate goal in \nf{FlipIt} is to control the resource for the maximum possible time while also maximizing its score. Without the second constraint, the best strategy would be to flip at each iteration to ensure control of the resource throughout the game. In this paper, we have seen that the defender is able to learn a strategy that maximizes its benefits when opposed to strategies with an expected interval time between two actions to be greater than the flip move cost. However, when opposed to highly active opponents (with an expected interval time between two moves smaller than the flip cost), the defender learns that a no-play strategy is the best strategy as it maximizes its overall score, and the opponent's excessive flips forces the defender to ``drop out'' of the game. This is an interesting behavior, one we would like to further exploit and integrate in a game where probabilistic moves or changes of strategy throughout the game would be possible. Moreover, there are many ways to expand \nf{FlipIt} to model real world situations and we intend on pursuing this project to analyze the use of reinforcement learning in different variants of the game. A few of our interests include games with probabilistic moves and team-based multiplayer games. Adding an upper bound on the budget and limiting the number of flips allowed per player would force players to flip more efficiently throughout the game, where the addition of \nf{check} could be key to developing new adaptive strategies. Finally, instead of having individual players competing against each other as shown previously, we would like to analyze a team-based variant of \nf{FlipIt} where players on a same team can either coordinate an attack on the device in question, or cooperate in order to defend the resource and prevent any intrusions.

\section{Conclusion}

Cyber and real-world security threats often present incomplete or imperfect state information. We believe our framework is well equipped to handle the noisy information and to learn an efficient counter-strategy against different classes of opponents for the game \nf{FlipIt}, even under partial observability, regardless of the number of opponents. Such strategies can be applied to optimally schedule key changes to network security amongst many other potential applications. Furthermore, we extended the game to larger action-spaced \nf{FlipIt}, where the additional lower cost action \nf{check} was introduced, allowing agents to obtain useful feedback regarding the current state of the game and to plan future moves accordingly. Finally, we made our source code publicly available for reproducibility purposes and to encourage researchers to further investigate adaptive strategies in \nf{FlipIt} and its variants.

\bibliographystyle{abbrv}
\bibliography{main}

\end{document}